\documentclass[10pt,twocolumn,letterpaper]{article}

\usepackage[pagenumbers]{cvpr}

\usepackage{makecell, graphicx, romannum, amssymb, setspace, algpseudocode, enumitem, pifont, etoolbox, fontawesome5, longtable, array, arydshln, bbding, multirow, caption, capt-of, booktabs, tabularx}
\usepackage[most]{tcolorbox}
\usepackage[ruled, lined, linesnumbered, commentsnumbered, longend]{algorithm2e}

\usepackage{amsmath,amsfonts,bm}

\def\eqref#1{equation~\ref{#1}}
\def\1{\bm{1}}

\DeclareMathAlphabet{\mathsfit}{\encodingdefault}{\sfdefault}{m}{sl}
\SetMathAlphabet{\mathsfit}{bold}{\encodingdefault}{\sfdefault}{bx}{n}

\definecolor{cvprblue}{rgb}{0.21,0.49,0.74}
\usepackage[pagebackref,breaklinks,colorlinks,allcolors=cvprblue]{hyperref}

\renewcommand{\paragraph}[1]{\vspace{1.25mm}\noindent\textbf{#1}}

\newtheorem{theorem}{Theorem}

\setcitestyle{square, comma}

\title{Look Every Frame All at Once: Video-Ma$^2$mba for Efficient Long-form Video Understanding with Multi-Axis Gradient Checkpointing}

\author{%
  Hosu Lee\thanks{Equal contribution. $\dagger$ Corresponding author.}~~~~~~~~~~Junho Kim\footnote[1]{}~~~~~~~~~~Hyunjun Kim~~~~~~~~~~Yong Man Ro\footnote[2]{}\\
Integrated Vision and Language Lab, KAIST, South Korea \\
  \tt\small{\{leehosu01,~arkimjh,~kimhj709,~ymro\}@kaist.ac.kr} \\
{\href{https://ivy-lvlm.github.io/Video-MA2MBA/}{https://ivy-lvlm.github.io/Video-MA2MBA}}
}

\begin{document}
\pagenumbering{arabic}
\maketitle
\begin{abstract}

With the growing scale and complexity of video data, efficiently processing long video sequences poses significant challenges due to the quadratic increase in memory and computational demands associated with existing transformer-based Large Multi-modal Models (LMMs). To address these issues, we introduce Video-Ma$^2$mba, a novel architecture that incorporates State Space Models (SSMs) within the Mamba-2 framework, replacing the attention mechanisms. This allows the LMMs to scale linearly in terms of time and memory requirements, making it feasible to handle long-duration video content. Furthermore, we enhance the memory efficiency introducing the Multi-Axis Gradient Checkpointing (MA-GC) method, which strategically manages memory by retaining only essential activations across multiple computational axes. Our approach significantly reduces the memory footprint compared to standard gradient checkpointing. Empirical analyses show that Video-Ma$^2$mba can process extensive video sequences\textemdash equivalent to millions of tokens or over two hours of continuous sequences at 1 FPS\textemdash on a single GPU. By maintaining a detailed capture of temporal dynamics, our model improves the accuracy and relevance of responses in long video understanding tasks, demonstrating substantial advantages over existing frameworks.

\end{abstract}    
\section{Introduction}
\label{sec:intro}

As video data grows in scale and complexity, the demand for models capable of efficiently processing long video sequences has intensified. Transformer-based models~\cite{brown2020language, touvron2023llama} have become central to sequence processing due to their effectiveness and versatility in handling complex dependencies as input frames increase. With the emergence of the Large Language Models (LLMs) era~\cite{chatgpt, gpt4, Gemini}, video understanding models have entered a new phase. Thanks to the core capabilities of zero-shot learning and strong reasoning in LLMs, various Large Multi-modal Models (LMMs)~\cite{dai2023instructblip, liu2023visual, liu2023improved} have achieved enhanced cross-modality consistency, particularly between vision and language. Accordingly, diverse video-LMMs~\cite{maaz2023video, lin2023video, li2024mvbench} have been proposed to understand video content by integrating spatio-temporal information into LLMs, and achieved comparable reasoning performances to comprehend complex visual narratives and temporal dynamics.

%%%%%%%%%%%%%%%%%%%%%%%%%%%%%%%%%%%%%%%%%%%%%%%%%%%%%%%%%%%%%%%%%%%%%%%%%%%%%%%%%%%%%%%%%%%%%%%%%%%%%%%%%%%
\begin{figure}[t]
    \centering
    \includegraphics[width=0.99\linewidth]{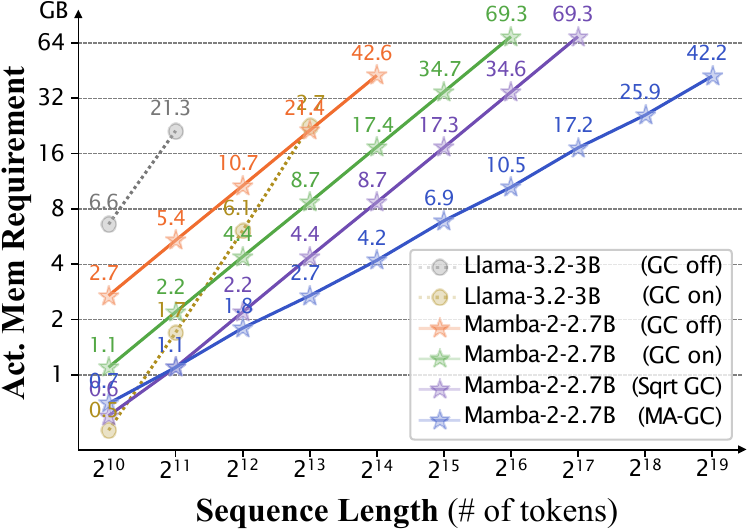}
    \vspace{-0.3cm}
    \caption{Memory usage comparison across sequence lengths for Mamba-2-2.7B with different checkpointing methods, demonstrating the memory-saving capability of Multi-Axis Gradient Checkpointing (MA-GC).}
    \vspace{-0.5cm}
    \label{fig:1}
\end{figure}
%%%%%%%%%%%%%%%%%%%%%%%%%%%%%%%%%%%%%%%%%%%%%%%%%%%%%%%%%%%%%%%%%%%%%%%%%%%%%%%%%%%%%%%%%%%%%%%%%%%%%%%%%%%

However, current video-LMMs face critical challenges when applied to longer video sequences integrated into the LLM structure, as the attention mechanism~\cite{Vaswani2017AttentionIA}\textemdash the core component of Transformers\textemdash incurs memory load and computational costs that scale quadratically with sequence length. This growth in resource requirements becomes prohibitive for processing extended video sequences, where sequence lengths often exceed 128K tokens (approximately 400 frames), resulting in significant inefficiencies in memory consumption and computational load.

To address the challenge for the long video understanding, various strives have been evident with several different approaches: (\lowercase\expandafter{\romannumeral1}) sparse and uniform sampling (\textit{e.g.,} typically $8$ or $16$ frames) to reduce the number of input video frames into LMMs, which is the most common method~\cite{lin2024vila, lin2023video, li2024llava} and (\lowercase\expandafter{\romannumeral2}) memory-augmented generation~\cite{he2024ma, song2024moviechat} that stores long-term visual information and later access to the memory bank. More recently, Zhang~\textit{et al.}~\cite{zhang2024long} have proposed a method for transferring long context by employing RoPE-based~\cite{Su2021RoFormerET} frequency extension in LM backbones within existing memory limits, which enables the models to process more visual tokens for the longer video sequences. Despite such efforts, the exponential growth in memory usage with increasing sequence lengths imposes fundamental limitations on the amount of information that can be fed into the model. Consequently, when the model is prompted to respond based on missing information (or frames), it often generates responses that are irrelevant and disconnected from the facts or user queries.

Here, the core difficulty lies in the quadratic time and space complexity of the Transformer's attention mechanism, which restricts efficient processing of lengthy video sequences. To enable more practical scalability when handling longer sequences, the fundamental solution is on shifting from quadratic complexity to linear complexity, facilitating more scalable processing for the video data. To do so, in this paper, we introduce \textbf{Video-Ma$^2$mba}, specifically designed to handle extremely long video sequences all at once. We substitute the Transformer-based LLMs~\cite{touvron2023llama, vicuna} to Mamba-2~\cite{Dao2024TransformersAS} structure, utilizing State Space Models (SSMs)~\cite{Gu2023MambaLS} as a replacement for the attention mechanism. This allows our framework to not only preserve the effectiveness of sequence processing but also enhance memory efficiency, thereby achieving linear time and space complexity with respect to the sequence length. 

In addition to the architectural changes, to push the boundary of memory utilization within Mamba-2 architecture, we present a new \textbf{Multi-axis Gradient Checkpointing} (MA-GC) method. Specifically, the Gradient checkpointing (GC)~\cite{Chen2016TrainingDN} strategically saves selected activations throughout the computational graph, allowing only partial activations to be re-computed during the backward pass (therefore, widely used for managing the substantial memory demands of extensive attention layers in Transformer-based LLMs). On the other hands, due to the nature of Mamba structure, which belongs partially observed Markov models where the hidden state progresses over time according to a Markov process rather than dense interaction models (\textit{e.g.,} Transformer), we can implement another axis for GC in the sequential direction by selectively retaining only those sequence-wise activations necessary for backpropagation. The MA-GC reduces memory usage from the $O(L \cdot S)$ complexity of the original Mamba-2 to $O(S)$ by applying GC in multiple directions, which allows Video-Ma$^2$mba to process full video sequences at 1-second intervals without the need for frame sampling. Our empirical analyses in \cref{fig:1} indicate that the proposed method performs effectively in handling extended sequences, successfully processing sequence lengths in the millions on a single GPU, corresponding to over 2-hours of continuous video input at a frame rate of 1 FPS. By observing each frame at regular intervals, our approach captures more comprehensive temporal information than uniform sampling, providing improvements in model responses and memory efficiency. 

Through extensive experiments and computational analyses on MA-GC, we demonstrate that Video-Ma$^2$mba efficiently manages resource demands, effectively breaking the quadratic memory growth for handling long sequence.

Our contribution can be summarized into three-fold:
\begin{itemize} 
\item We propose Video-Ma$^2$mba, a new multi-modal framework designed to handle extensively long video sequences without losing frame information, by replacing the Transformers with Mamba-2 architecture.

\item To significantly enhance memory utilization within our framework, we introduce the Multi-axis Gradient Checkpointing strategy. Our strategy selectively stores activations in a bi-axis direction, effectively reducing the space complexity to $O(S)$.

\item Through extensive evaluation and analyses, we corroborate that our framework can efficiently process sequence lengths in the millions, corresponding to up to 2-hours of video sequence at 1 FPS with competent performance.
\end{itemize}

\section{Related Work}
\label{sec:related}
\subsection{Context Extension Methods}

Training models with extended sequence lengths has become increasingly challenging due to the computational and memory demands associated with scaling sequence lengths. The standard Transformer models~\cite{Vaswani2017AttentionIA, brown2020language} struggle with the quadratic complexity of attention mechanisms, which quickly becomes infeasible for long sequences. To address these limitations, various methods have been proposed to extend the context length either during or after pre-training.

One of the common approaches is to generalize knowledge learned over shorter sequences to longer sequences~\cite{Dai2019TransformerXLAL, Press2021TrainST}. However, this approach can lead to issues in positional extrapolation, as positional encodings trained on shorter contexts may not generalize well to longer contexts. Techniques such as Rotary Position Embedding~\cite{Su2021RoFormerET} and Position Interpolation~\cite{Chen2023ExtendingCW} mitigate this by modifying the positional embedding, making it better suited for extrapolation beyond the training range. Additionally, ALiBi \cite{Press2021TrainST} has applied attention biases directly, improving context extension without a strict reliance on positional encodings.

An alternative approach is leveraging Structured State Space Models (SSMs)~\cite{Gu2023MambaLS}, which inherently achieve linear time and space complexity with respect to the sequence length. This efficiency stems from the intrinsic properties of SSMs, allowing for effective training on extended contexts and facilitating long-sequence learning without the prohibitive memory costs.

\subsection{Long Video Understanding with LMMs}

Long video understanding presents unique challenges due to the need to capture dependencies across extended sequences while managing high memory consumption. The core challenges in handling long videos are memory limitations and finite context lengths. Several studies~\cite{zhang2024beyond, jin2024chat} have addressed this by sparsely sampling video frames (typically 8 or 16 from the video instance rather than using dense fps-based sampling), or by employing token compression to reduce data to a more manageable size~\cite{maaz2023video, li2025llama}. Additionally, memory-augmented approaches~\cite{he2024ma, song2024moviechat} have been proposed to store relevant information beforehand and recall explicit knowledge when generating responses.

While such methods are simpler to implement and effective for managing memory, they risk missing critical details in long video content. Our approach addresses these limitations by utilizing full-frame sequences at 1 FPS, ensuring comprehensive temporal representation without relying on sparse sampling and achieving memory efficiency.

\subsection{Gradient Checkpointing Techniques}
The gradient checkpointing (GC) is a well-established method for reducing memory usage in deep learning models by selectively storing intermediate activations and recomputing them as needed during the backward pass. Initially developed to manage memory constraints in training, this approach allows models to trade off additional computation for reduced memory requirements.

Chen \textit{et al.}~\cite{Chen2016TrainingDN} have introduced fundamental checkpointing techniques applicable across deep networks and recurrent neural networks (RNNs). For deep networks, the technique involves segmenting the network along the layer axis, storing the outputs at segment boundaries, and recomputing the intermediate results within each segment as needed. This segmentation reduces memory requirements from $O(n)$ to $O(\sqrt{n})$, where $n$ is the number of layers. For RNNs, a similar approach is used along the time-axis, allowing memory usage to scale sublinearly with sequence length by storing checkpoints at specific time intervals.

Our work builds on these principles by applying the GC in both the layer and sequence dimensions, enabling efficient processing of long sequences across bi-directional axes. Our approach enhances memory efficiency and is particularly well-suited for understanding long videos, where integrating both temporal and spatial contexts is crucial. 
\section{Video-Ma$^2$mba}
\label{sec:Method}

\paragraph{Overview.} We first elaborate on the distinction of the Mamba-2 architecture in handling memory efficiency, then introduce a new gradient checkpointing method that can be a key factor in extending the sequence length of Mamba-2. By seamlessly implementing our new context extension strategy during the training of Video-Ma$^2$mba, our video model can handle up to maximum $0.8$M input sequence tokens and overcome current challenges in long video understanding relying on partial frame sampling.

\subsection{Preliminary: Mamba-2 and Simplification}

The Mamba model~\cite{Gu2023MambaLS} initially has leveraged structured state space models (SSMs) to efficiently handle sequence modeling. Building on this foundation, Mamba-2~\cite{Dao2024TransformersAS} represents a further advancement to scale up to larger state sizes with the concept of Structured State-Space Duality (SSD), which enhances sequence processing capabilities through time-varying state transitions and input-output mappings, thus more effective handling for sequence data. The general form of SSD in Mamba-2 can be formulated as:
%%%%%%%%%%%%%%%%%%%%%%%%%%%%%%%%%%%%%%%%%%%%%%%%%%%%%%%%
\begin{equation}
    h_t = A_t h_{t-1} + B_t x_t, \quad y_t = C_t h_t,
\label{eqn:mamba}
\end{equation}
%%%%%%%%%%%%%%%%%%%%%%%%%%%%%%%%%%%%%%%%%%%%%%%%%%%%%%%%
where $A_t \in \mathbb{R}^{N \times N}$, $B_t \in \mathbb{R}^{N \times 1}$, and $C_t \in \mathbb{R}^{1 \times N}$ are state matrices that vary over time, allowing the model to adapt dynamically to different input structures. This time variance, or \textit{selectivity}, enhances Mamba-2's flexibility in comparison to linear time-invariant SSMs. By employing time-varying matrices $A_t$, $B_t$, and $C_t$, Mamba-2 can be seen as a selective SSM that performs sequential updates to the hidden state $h_t$ based on previous states and current inputs. 

This selective structure is particularly advantageous in capturing sequence dynamics over longer frames, which standard Recurrent Neural Networks (RNNs)~\cite{Elman1990FindingSI} with fixed parameters struggle to achieve. At the same time, Mamba-2 also shares some similarities with a certain RNN framework when non-linear activations are removed. A standard RNN with a non-linear activation function $\sigma$ (\textit{e.g.}, Tanh or ReLU) updates its hidden state as follows:
%%%%%%%%%%%%%%%%%%%%%%%%%%%%%%%%%%%%%%%%%%%%%%%%%%%%%%%%
\begin{equation}
    h_t = \sigma(A h_{t-1} + B x_t), \quad y_t=\sigma(Ch_t).
\label{eqn:mamba2}
\end{equation}
%%%%%%%%%%%%%%%%%%%%%%%%%%%%%%%%%%%%%%%%%%%%%%%%%%%%%%%%
Here, removing the activation function $\sigma$ transforms the RNN, making its structure similar to that of SSD:
%%%%%%%%%%%%%%%%%%%%%%%%%%%%%%%%%%%%%%%%%%%%%%%%%%%%%%%%
\begin{equation}
    h_t = A h_{t-1} + B x_t, \quad y_t = C h_t.
\label{eqn:mamba3}
\end{equation}
%%%%%%%%%%%%%%%%%%%%%%%%%%%%%%%%%%%%%%%%%%%%%%%%%%%%%%%%

Consequently, SSD can be regarded as a simplified version of RNN, where the time-varying parameters of SSD introduce a level of flexibility and adaptability that fixed-parameter RNNs lack. This dynamic modification allows SSD to effectively address challenges associated with static parameter models in handling complex temporal sequences.

\subsection{Multi-Axis Gradient Checkpointing}

\begin{figure}[t!]
    \centering
    \includegraphics[width=0.99\linewidth]{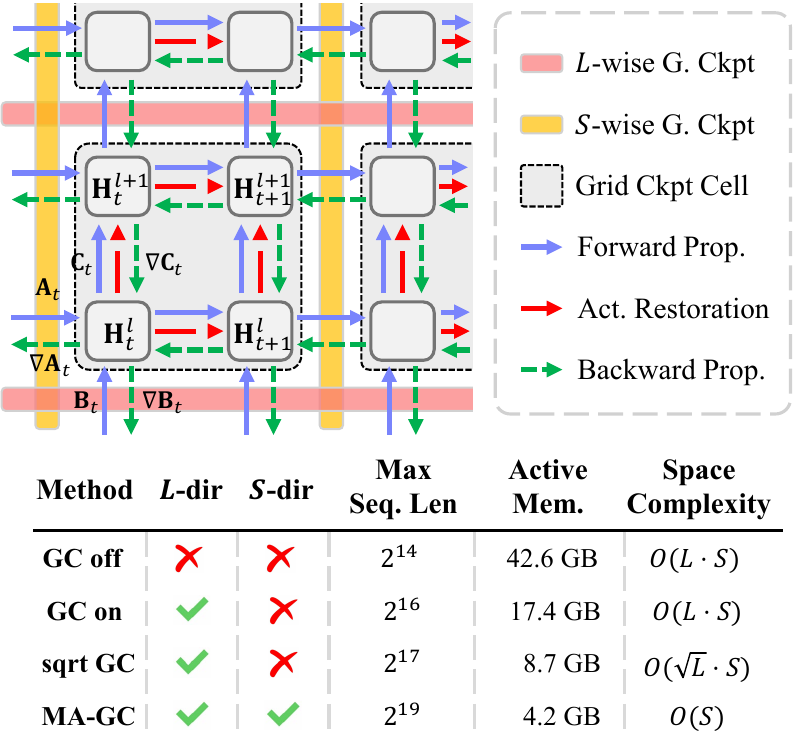}
    \vspace{-0.3cm}
    \caption{Overview of MA-GC grid structure. Checkpoints are stored every $l$ layers and $s$ steps. The \textcolor[HTML]{7984FC}{blue}, \textcolor[HTML]{FF0000}{red}, and \textcolor[HTML]{00B050}{green} arrows indicate forward propagation, activation restoration, and gradient propagation, respectively. This grid design optimizes memory by selectively restoring activations as needed. The below table shows comparison of checkpointing usage, maximum sequence length on 80GB VRAM, and peak activation memory in BFloat16 at sequence length 16384.}
    \label{fig:2}
    \vspace{-0.3cm}
\end{figure}

Considering that Mamba-2 follows RNN-like structure, as illustrated in \cref{fig:2}, when processing the SSD state $H^l_{i+1}$, it only requires the prior state $H^l_{i}$, not $H^l_{i-1}$, unlike Transformers that require all previous states to calculate attention weights across the entire input sequence. Here, it is important to note that this distinction enables us to introduce an additional gradient checkpointing axis, not only along the layer direction but also uniquely along the sequence direction, which attribute to the architectural properties of Mamba-2 (whereas Transformer cannot achieve).

Our key motivation for employing bi-axis checkpointing lies in its effectiveness at managing memory demands, which enables the processing of extremely long video sequences in their entirety without needing to sample scenes partially. Here, we introduce a new GC strategy, \textbf{Multi-Axis Gradient Checkpointing} (MA-GC), that not only increases the feasible sequence length up to $2^{19}$ but also substantially cuts activation memory usage. While previous methods such as the $\sqrt{L}$ layer grouping~\cite{Chen2016TrainingDN} achieved some memory reduction by applying checkpoints every $\sqrt{L}$ layers, they were inadequate for very long sequences. In contrast, our MA-GC method applies checkpointing along both layer and sequence axes, reducing space complexity from $O(\sqrt{L} \cdot S)$ in standard GC to just $O(S)$. This significant improvement allows our model to process longer sequences more efficiently without partial frame sampling, thus supporting extended sequence lengths in understanding long video content.

Specifically, as shown in \cref{fig:2}, our MA-GC strategy involves two checkpoint types in the forward pass for the given $S$ sequence length and $L$ stacked layers: (\lowercase\expandafter{\romannumeral1}) Layer-wise checkpoints, where layer activations are stored every $l$ layers, and (\lowercase\expandafter{\romannumeral2}) Sequence-wise checkpoints, where states across all layers are stored every $s$ time steps. The intersecting points of these two checkpoints create \textit{grid cells} that are essential for efficient backpropagation. Within each grid cell, activations are sequentially restored and gradients are propagated in an efficient manner. This grid-based structure facilitates selective reconstruction of states only when necessary, thereby optimizing memory usage during the computationally intensive backpropagation process. We provide detailed explanations of both the forward and backward processes in \cref{alg:ma_gc_forward} and \cref{alg:ma_gc_backward}.

%%%%%%%%%%%%%%%%%%%%%%%%%%%%%%%%%%%%%%%%%%%%%%%%%%%%%%%%%%%%%%%%%%%%%%%%%%%%%%%%
%%%%%%%%%%%%%%%%%%%%%%%%%%%%%%%%%%%%%%%%%%%%%%%%%%%%%%%%%%%%%%%%%%%%%%%%%%%%%%%%

\subsection{Analysis of Upper Bound of Memory Reduction}
\label{sec:analysis}

To understand how MA-GC reduces memory usage, we analyze its space complexity and establish an upper bound on memory savings. In a naive RNN (or similarly, in Mamba or Mamba-2) architecture with $L$ layers and a sequence length $S$, backpropagation requires storing activation memory of $\Theta(L \cdot S)$ due to the need to store activations for each layer and each time step during the backward pass.

In our proposed MA-GC method, the required memory $M$ is given by:
%%%%%
\begin{equation}
M = \underbrace{\frac{L S}{l}}_{\text{L-wise G. Ckpt}} + \underbrace{\frac{L S}{s}}_{\text{S-wise G. Ckpt}} + \underbrace{l s}_{\text{Grid Ckpt Cell}},
\label{eq:MemoryEquation}
\end{equation}
%%%%%
where $l$ and $s$ represent the checkpoint intervals along the layer and sequence directions, respectively. The first term accounts for the memory needed for layer-wise checkpoints, the second term for sequence-wise checkpoints, and the third term for activations stored within each grid cell during recomputation, respectively.

%%%%%%%%%%%%%%%%%%%%%%%%%
\begin{algorithm}[t!]
    \SetAlgoLined
    \SetKwInOut{KwIn}{Input}
    \SetKwInOut{KwOut}{Output}
    
    \KwIn{Input sequence $\{x_i\}_{i=1}^S$, Total layers $L$, Intervals $l_{\text{int}}$, $s_{\text{int}}$}
    \KwOut{Checkpoints $L_{\text{ckpt}}$, $S_{\text{ckpt}}$}
    
    \textbf{Initialization:} $L_{\text{ckpt}} \leftarrow \texttt{defaultdict}(\texttt{list})$, $S_{\text{ckpt}} \leftarrow \texttt{defaultdict}(\texttt{list})$\;
    
    \For{$i \leftarrow 1$ \KwTo $S$}{
         $x \leftarrow x_i$\;

         $h \leftarrow \mathbf{0}$\;
        
        \For{$j \leftarrow 1$ \KwTo $L$}{
            \If{$j \bmod l_{\text{int}} = 0$}{
                Append $x$ to $L_{\text{ckpt}}[\lceil i / s_{\text{int}} \rceil, \lceil j / l_{\text{int}} \rceil]$\;
            }
            \If{$i \bmod s_{\text{int}} = 0$}{
                Append $h$ to $S_{\text{ckpt}}[\lceil i / s_{\text{int}} \rceil, \lceil j / l_{\text{int}} \rceil]$\;
            }
            $(x, h) \leftarrow \text{Layer}_j(x, h)$\;
        }
    }
    \KwRet{$L_{\text{ckpt}}$, $\text{S}_{\text{ckpt}}$}
    \caption{Checkpointing Strategy with MA-GC (Forward)}
    \label{alg:ma_gc_forward}
\end{algorithm}

%%%%%%%%%%%%%%%%%%%%%%%%%

Our objective is to minimize $M$ by finding optimal values for $l$ and $s$, thus maximizing memory savings. To achieve this, we employ the \textit{Extreme Value Theorem} and \textit{Fermat's Theorem} to identify the minimum values of a differentiable function on a closed interval.
%%%%%%%%%%%%%%%%%%%%%%%%%%%%%%%%%%%%%%%%%%%%%
\begin{theorem}[Extreme Value Theorem]
\label{thm:extreme_value}
If $f$ is continuous on a closed interval $[a, b]$, then $f$ achieves both maximum and minimum values on $[a, b]$.
\end{theorem}
%%%%%%%%%%%%%%%%%%%%%%%%%%%%%%%%%%%%%%%%%%%%%
\begin{theorem}[Fermat's Theorem]
\label{thm:fermat}
If $f$ is differentiable on an open interval $(a, b)$ and has a local extremum at an interior point $c \in (a, b)$, then $f'(c){=}0$.
\end{theorem}
%%%%%%%%%%%%%%%%%%%%%%%%%%%%%%%%%%%%%%%%%%%%%
According to these theorems, the minimum value of $M(l, s)$ occurs either at a critical point (where $\frac{\partial M}{\partial l} {=} 0$ and $\frac{\partial M}{\partial s} {=} 0$) or at the boundary of the region defined by $1 \leq l \leq L$ and $1 \leq s \leq S$. First, we calculate the partial derivatives of $M$:
%%%%%%%%%%%%%%%%%%%%%%%%%%%%%%%%%%%%%%%%%%%%%
\begin{equation}
\frac{\partial M}{\partial l} = -\frac{L S}{l^2} + s = 0,
\label{eq:PartialDerivativeL}
\end{equation}
%%%%%%%%%%%%%%%%%%%%%%%%%%%%%%%%%%%%%%%%%%%%%
\begin{equation}
\frac{\partial M}{\partial s} = -\frac{L S}{s^2} + l = 0.
\label{eq:PartialDerivativeS}
\end{equation}
%%%%%%%%%%%%%%%%%%%%%%%%%%%%%%%%%%%%%%%%%%%%%
Solving these, we find that the critical point occurs at $l {=} s {=} \sqrt[3]{L S}$, with the corresponding minimal memory requirement $M^*_{\text{critical}}$:
%%%%%%%%%%%%%%%%%%%%%%%%%%%%%%%%%%%%%%%%%%%%%
\begin{equation}
M^*_{\text{critical}} = 3 (L S)^{\frac{2}{3}}.
\label{eq:MemoryCritical}
\end{equation}
%%%%%%%%%%%%%%%%%%%%%%%%%%%%%%%%%%%%%%%%%%%%%
However, since $l$ and $s$ are integers within $1 \leq l \leq L$ and $1 \leq s \leq S$, we should evaluate boundary cases. To do so, we use the \textbf{Arithmetic Mean-Geometric Mean (AM-GM) inequality} to derive upper bounds for the minimum values of $l$ and $s$. The AM-GM inequality states that for non-negative real numbers $a$ and $b$, their arithmetic mean is at least their geometric mean: $\frac{a + b}{2} \geq \sqrt{a b}$. We can express lower bound of each bound as:
%%%%%%%%%%%%%%%%%%%%%%%%%%%%%%%%%%%%%%%%%%%%%
\begin{equation}
M^*_{l\text{-bound}} = S + L \left(\frac{S}{s} + s\right) \geq S + 2L\sqrt{S},
\label{eq:MemoryLBound}\end{equation}
%%%%%%%%%%%%%%%%%%%%%%%%%%%%%%%%%%%%%%%%%%%%%
\begin{equation}
M^*_{s\text{-bound}} = L + S \left(\frac{L}{l} + l\right) \geq L + 2S\sqrt{L}.
\label{eq:MemorySBound}\end{equation}
%%%%%%%%%%%%%%%%%%%%%%%%%%%%%%%%%%%%%%%%%%%%%
Then, the overall optimized memory $M^*$ is selected based on the regions where each configuration achieves balance according to the AM-GM inequality, as follows:
%%%%%%%%%%%%%%%%%%%%%%%%%%%%%%%%%%%%%%%%%%%%%
\begin{equation}
M^* =
\begin{cases}
\Theta\big((L S)^{\frac{2}{3}}\big) & \text{if } L \leq S^{2} \text{ and } S \leq L^{2}, \\
\Theta(S) & \text{if } L^{2} \leq S, \\
\Theta(L) & \text{if } S^{2} \leq L.
\end{cases}
\label{eq:MemoryCases}\end{equation}
%%%%%%%%%%%%%%%%%%%%%%%%%%%%%%%%%%%%%%%%%%%%%
Since we are interested in training longer sequences with a fixed number of layers (\textit{i.e.}, $L < S$), we focus on the region where $L^2 \leq S$. Thus, the memory required simplifies to $M^* {=} \Theta(S)$, and the memory savings ratio $\frac{L S}{M^*}$ is given by:
%%%%%%%%%%%%%%%%%%%%%%%%%%%%%%%%%%%%%%%%%%%%%
\begin{equation}
\frac{L S}{M^*} = \Theta(L).
\label{eq:MemorySavingsRatio}
\end{equation}
%%%%%%%%%%%%%%%%%%%%%%%%%%%%%%%%%%%%%%%%%%%%%
Thus, as the sequence length $S$ grows, the upper bound on memory savings achieved by MA-GC scales proportionally to the number of layers $L$. This analysis demonstrates that MA-GC effectively reduces space complexity from $\Theta(L S)$ to $\Theta(S)$, enabling the processing of very long sequences with significantly reduced memory constraints.

As shown in \cref{fig:1}, our experimental results validate this theoretical analysis. Without gradient checkpointing, the activation memory requirement at sequence length $2^{14}$ is $42.6$ GB, whereas with MA-GC, only $42.2$ GB is required at sequence length $2^{19}$. This demonstrates the practical effectiveness of MA-GC in drastically reducing memory usage while enabling to handle extremely long sequences.

%%%%%%%%%%%%%%%%%%%%%%%%%%%%%%%%%%%%%%%%%%%%%

\begin{algorithm}[t!]
    \SetAlgoLined
    \SetKwInOut{KwIn}{Input}
    \SetKwInOut{KwOut}{Output}
    
    \KwIn{Gradients $\nabla_y$, $\nabla_{\text{state}}$, Checkpoints $L_{\text{ckpt}}$, $S_{\text{ckpt}}$, Total layers $L$, Intervals $l_{\text{int}}$, $s_{\text{int}}$}
    
    \textbf{Initialization:} $\nabla_h[j] \leftarrow \nabla_{\text{state}}[(j-1) \cdot l_{\text{int}} : j \cdot l_{\text{int}}]$ for each $j = 1, \dots, \lceil L / l_{\text{int}} \rceil$\;

    \For{$i_{\text{blk}} \gets S$ \KwTo $1$ \textnormal{\textbf{by}} $-s_{\text{int}}$}{
        $\nabla_x \leftarrow \nabla_y[i_{\text{blk}} - s_{\text{int}} : i_{\text{blk}}]$

        \For{$j_{\text{blk}} \gets L$ \KwTo $1$ \textnormal{\textbf{by}} $-l_{\text{int}}$}{
            $i_{\text{idx}} \leftarrow \lceil i_{\text{blk}} / s_{\text{int}} \rceil $\;
            
            $j_{\text{idx}} \leftarrow \lceil j_{\text{blk}} / l_{\text{int}} \rceil $\;

            $x_{\text{ckpt}} \leftarrow L_{\text{ckpt}}[i_{\text{idx}}, j_{\text{idx}}]$\;

            $h_{\text{ckpt}} \leftarrow S_{\text{ckpt}}[i_{\text{idx}}, j_{\text{idx}}]$\;
            
            $(x, h) \leftarrow \text{Recompute\_Forward}(x_{\text{ckpt}}, h_{\text{ckpt}}, i_{\text{blk}}, j_{\text{blk}})$\;

            $(\nabla_x, \nabla_h[j_{\text{idx}}]) \leftarrow \text{Backward}(x, h, \nabla_x, \nabla_h[j_{\text{idx}}])$\;
        }
    }
    \caption{Backpropagation with Grid-Cell Restoration on MA-GC (Backward)}
    \label{alg:ma_gc_backward}
\end{algorithm}

%%%%%%%%%%%%%%%%%%%%%%%%%%%%%%%%%%%%%%%%%%%%%

\subsection{Model Architecture}

\begin{figure}[t]
    \centering
    \includegraphics[width=0.99\linewidth]{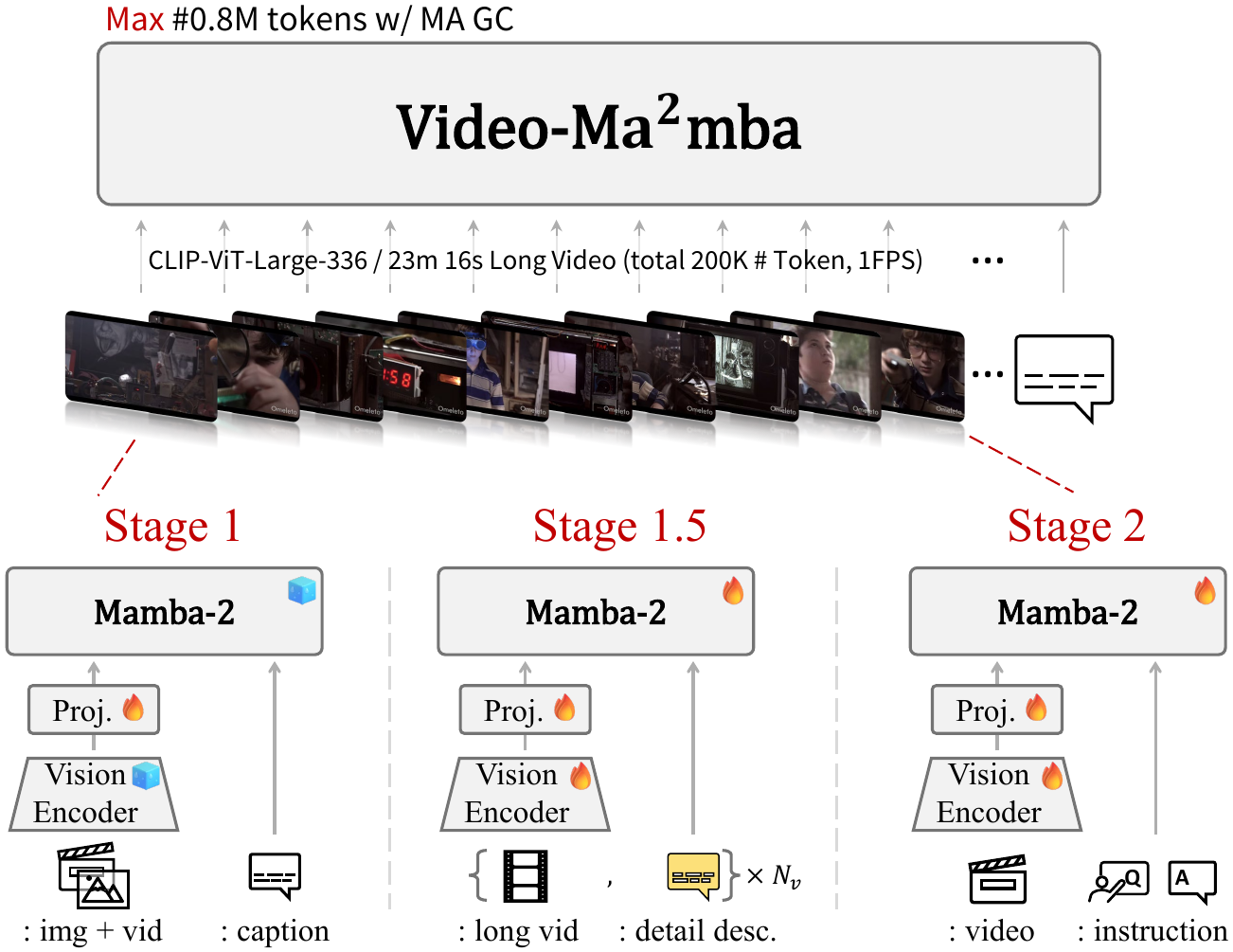}
    \vspace{-0.4cm}
    \caption{The overall summarization for the training stages of Video-Ma$^2$mba.}
    \vspace{-0.5cm}
    \label{fig:3}
\end{figure}

Now, we move on to training Video-Ma$^2$mba with long video data using the proposed MA-GC strategy. Analogous to the widely adopted LMM architecture~\cite{liu2023visual, liu2023improved}, as outlined in \cref{fig:3}, Video-Ma$^2$mba follows three main structural components: (\lowercase\expandafter{\romannumeral1}) vision encoder to process input video frames, (\lowercase\expandafter{\romannumeral2}) cross-modal projector to align vision-text modalities, and (\lowercase\expandafter{\romannumeral3}) LLM backbone with Mamba-2 architecture (370M / 1.3B / 2.7B). 

To ensure thorough coverage of the video content, we systematically sample each frame at 1 FPS. We employed \texttt{CLIP-ViT-L-336px}~\cite{radford2021learning} as our vision encoder, extracting features from the penultimate layer, followed by 2x2 bilinear pooling to produce $144$ visual tokens per frame. Then, we transform these features to align with the language model’s embedding space using a lightweight 2-layer MLP projection with GELU activation. For the language model backbone, we use the Mamba-2 structure instead of standard Transformer-based LLMs, enabling more efficient embedding process with newly designed MA-GC strategy.

%%%%%%%%%%%%%%%%%%%%%%%%%%%%%%%%%%%%%%%%%%%%%%%%%%%%%%%%%%%%%%%%%%%%%%%%%%%%%%%%
%%%%%%%%%%%%%%%%%%%%%%%%%%%%%%%%%%%%%%%%%%%%%%%%%%%%%%%%%%%%%%%%%%%%%%%%%%%%%%%%

\subsection{Training Stages} Our training pipeline consists of three stages as summarized in Fig.~\ref{fig:3}. Beyond the conventional two step training steps for LMMs: alignment training $+$ supervised fine-tuning with instruction data, Li \textit{et al.}~\cite{liu2024llavanext} have highlighted the importance of high-quality knowledge acquisition between these two stages (thus, stage 1.5). By exploiting the extended context length with the MA-GC during training, we reemphasize that our primary goal in training Video-Ma$^2$mba is to enhance the model’s ability to process and learn from long-form video data effectively. Here, expanding the stage 1.5 learning~\cite{liu2024llavanext}, we train Video-Ma$^2$mba using an interleaved learning approach with long video data~\cite{kim2024salova}, which comprises densely captioned video-text pairs covering entire long-sequence videos, each segment described in detail. By doing so, Video-Ma$^2$mba can preserve the narrative flow across video segments, enhancing its temporal understanding by learning the sequential relationships within the video content. Through the refined training approach, we aim to develop a robust capability in Video-Ma$^2$mba to handle complex, long-form video data in a contextually aware manner. 

\paragraph{Stage 1: Cross-modal Alignment.} During the initial step of training our framework, we utilize a dataset comprising a total of 790K image and video pairs with associated texts to ensure cross-modality consistency: 558K image-text pairs, filtered by LLaVA~\cite{liu2023visual}, and the remaining video-text pairs sampled from WebVid-2.5M~\cite{bain2021frozen}. During this stage, we only optimize the parameters in the projector layer.

\paragraph{Stage 1.5: Long Video Knowledge Learning.} As the intermediate long video parametric knowledge learning step, we utilize the SceneWalk dataset~\cite{kim2024salova} that consists of 11.8Khrs of YouTube videos from diverse categories (total 87.8K videos). This dataset includes $1.3$M segmented video clips, each with a corresponding scene-level detailed description. We train our model on the next-word generation task by applying an interleaved format to the segmented videos and their corresponding dense captions within each long video instance. In this stage, we unfreeze all parameters to facilitate comprehensive long video understanding.

\paragraph{Stage 2: Supervised Fine-Tuning.} We fine-tune our model on a diverse video QA dataset, including LLaVA-Video-178K~\cite{zhang2024video}, NeXT-QA~\cite{xiao2021next}, ActivityNetQA~\cite{yu2019activitynet}, and PerceptionTest~\cite{patraucean2024perception}. Together, these sources provide a total of 1.3 million video-instruction QA data~\textemdash caption entries, open-ended QA, and multiple-choice QA. During this stage, we fine-tune all parameters by unfreezing every network component to enhance the model's QA capabilities.
\section{Experiments}
\label{sec:Experiment}

\subsection{Experimental Setup}

\paragraph{Training Details.} To train our models, we used 1 node of 8 NVIDIA A100 GPUs, each with 80GB of memory. A cosine learning rate schedule was employed, with a learning rate of \(1 \times 10^{-3}\) for Stage 1 and \(4 \times 10^{-5}\) for both Stage 1.5 and Stage 2. We trained our models for one epoch at each step, with the entire training process taking approximately 4 days to complete for the 3.1B size model. All stages utilized BF16 precision, and we did not employ any form of parameter-efficient fine-tuning such as LoRA~\cite{hu2021lora}. To reduce the memory overhead from optimizer states, we utilized ZeRO-1 optimization~\cite{rajbhandari2020zero}, enabling more efficient memory management during training. Additional hyper-parameter details are provided in~\cref{sec:TrainHyperParameters}.

\paragraph{Implementation Details.} Our model, Video-Ma$^2$mba, uses \texttt{CLIP-ViT-L-336px} ($\approx$ 0.4B params) vision encoder paired with backbone LMs (370M / 1.3B / 2.7B), resulting in total model sizes of approximately 0.7B, 1.8B, and 3.1B, respectively. For the largest model configuration (3.1B), the backbone Mamba2-2.7B configuration specifically uses an embedding dimension of $2560$, $64$ layers, and a vocabulary size of $50,277$. The proposed MA-GC is applied throughout all training stages to maximize memory efficiency, enabling our model to handle a maximum sequence length of $0.8$M (approx. 1.5-hrs) during training, and generate responses with input sequences up to $2$M (approx. 4-hrs) with an 80GB VRAM size GPU.

\begin{table}[t!]
\centering
\begin{minipage}[t]{\linewidth}
    \centering
\resizebox{\linewidth}{!}{
\begin{tabular}{lccccc}
\Xhline{3\arrayrulewidth} \rule{0pt}{9pt}
                              &                                               & \multicolumn{4}{c}{Video-MME} \\ \cmidrule(lr){3-6}
\multirow{-2}{*}{Model}       & \multirow{-2}{*}{Size}                        & Short  & Medium & Long & Avg. \\ \hline
\rowcolor[HTML]{EFEFEF} 
GPT-4V~\cite{gpt4v}                        & \multicolumn{1}{c}{\cellcolor[HTML]{EFEFEF}-} & 70.5   & 55.8   & 53.5 & 59.9 \\
\rowcolor[HTML]{EFEFEF} 
GPT-4o~\cite{gpt4o}                         & \multicolumn{1}{c}{\cellcolor[HTML]{EFEFEF}-} & 80.0   & 70.3   & 65.3 & 71.9 \\
\rowcolor[HTML]{EFEFEF} 
Gemini 1.5 Pro~\cite{reid2024gemini}                & \multicolumn{1}{c}{\cellcolor[HTML]{EFEFEF}-} & 81.7   & 74.3   & 67.4 & 75.0 \\ \cdashline{1-6}\noalign{\vskip 0.5ex}
\rowcolor[HTML]{FFFFFF} 
ST-LLM~\cite{liu2025st}                        & 7B                                            & 45.7   & 36.8   & 31.3 & 37.9 \\
\rowcolor[HTML]{FFFFFF} 
VideoChat2-Mistral~\cite{li2024mvbench}            & 7B                                            & 48.3   & 37.0   & 33.2 & 39.5 \\
\rowcolor[HTML]{FFFFFF} 
Video-LLaVA~\cite{lin2023video}                   & 7B                                            & 45.3   & 38.0   & 36.2 & 39.9 \\
\rowcolor[HTML]{FFFFFF} 
ShareGPT4Video~\cite{chen2024sharegpt4video}                & 8B                                            & 48.3   & 36.3   & 35.0 & 39.9 \\
\rowcolor[HTML]{FFFFFF} 
Chat-UniVi-V1.5~\cite{jin2024chat}               & 7B                                            & 45.7   & 40.3   & 35.8 & 40.6 \\
\rowcolor[HTML]{FFFFFF} 
Qwen-VL-Chat~\cite{bai2023qwen}                  & 7B                                            & 46.9   & 38.7   & \underline{37.8} & 41.1 \\
\rowcolor[HTML]{FFFFFF} 
SliME~\cite{zhang2024beyond}                         & 8B                                            & \underline{53.3}   & \textbf{42.7}   & \textbf{39.8} & \textbf{45.3} \\ \cdashline{1-6}\noalign{\vskip 0.5ex}
\rowcolor[HTML]{ECF4FF} 
Video-Ma$^2$mba-0.7B & 0.7B                                          & 37.4   & 35.0   & 26.8 & 33.1 \\
\rowcolor[HTML]{ECF4FF} 
Video-Ma$^2$mba-1.8B & 1.8B                                          & 49.4   & 39.2   & 31.9 & 40.3 \\
\rowcolor[HTML]{ECF4FF} 
Video-Ma$^2$mba-3.1B   & 3.1B                                          & \textbf{57.6}   & \textbf{42.7}   & 35.4 & \underline{45.2} \\
\Xhline{3\arrayrulewidth} \rule{0pt}{9pt}
\end{tabular}
}
\vspace{-7.5mm}
\caption*{(a) Experimental results on Video-MME}

\end{minipage}
\begin{minipage}[t]{\linewidth}
    \centering
\resizebox{1.0\linewidth}{!}{
\begin{tabular}{lccccccc}
\Xhline{3\arrayrulewidth} \rule{0pt}{9pt}
                              &                        & \multicolumn{6}{c}{LongVideoBench}                  \\ \cmidrule(lr){3-8}
\multirow{-2}{*}{Model}       & \multirow{-2}{*}{Size} & \rotatebox{90}{8-15s} & \rotatebox{90}{15-60s} & \rotatebox{90}{180-600s} & \rotatebox{90}{900-3600s} & \rotatebox{90}{test set} & \rotatebox{90}{val set}  \\ \hline
\rowcolor[HTML]{EFEFEF} 
GPT-4o~\cite{gpt4o}                        & -                      & 71.6  & 76.8   & 66.7     & 61.6      & 66.7 & 66.7 \\
\rowcolor[HTML]{EFEFEF} 
Gemini 1.5 Pro~\cite{reid2024gemini}                & -                      & 70.2  & 75.3   & 65.0     & 59.1      & 64.4 & 64.0 \\ 
\rowcolor[HTML]{EFEFEF} 
GPT-4-Turbo~\cite{gpt4}                   & -                      & 66.4  & 71.1   & 61.7     & 54.5      & 60.7 & 59.1 \\ \cdashline{1-8}\noalign{\vskip 0.5ex}
VideoChat2~\cite{li2024mvbench}                    & 7B                     & 38.1  & 40.5   & 33.5     & 33.6      & 35.1 & 36.0 \\
VideoLLaVA~\cite{lin2023video}                    & 8B                     & 43.1  & 44.6   & 36.4     & 34.4      & 37.6 & 39.1 \\
PLLaVA~\cite{xu2024pllava}                        & 7B                     & 45.3  & 47.3   & 38.5     & 35.2      & 39.2 & 40.2 \\
LLaVA-1.5~\cite{liu2023improved}                     & 7B                     & 45.0  & 47.4   & \underline{40.1}     & 37.0      & 40.4 & \underline{40.3} \\
ShareGPT4Video~\cite{chen2024sharegpt4video}                & 7B                     & 46.9  & \underline{50.1}   & 40.0     & \textbf{38.7}      & \underline{41.8} & 39.7 \\ \cdashline{1-8}\noalign{\vskip 0.5ex}
\rowcolor[HTML]{ECF4FF} 
Video-Ma$^2$mba-0.7B & 0.7B                   & 43.3  & 45.4   & 33.3     & 28.5      & 34.2 & 34.0 \\
\rowcolor[HTML]{ECF4FF} 
Video-Ma$^2$mba-1.8B & 1.8B                   & \underline{48.4}  & 49.5   & 39.6     & 34.1      & 39.8 & 38.0 \\
\rowcolor[HTML]{ECF4FF} 
Video-Ma$^2$mba-3.1B   & 3.1B                   & \textbf{55.4}  & \textbf{55.6}   & \textbf{42.4}     & \underline{38.5}      & \textbf{44.2} & \textbf{43.0} \\
\Xhline{3\arrayrulewidth} \rule{0pt}{9pt}
\end{tabular}
}
\vspace{-7mm}
\caption*{(b) Experimental results on LongVideoBench}
\vspace*{-3mm}
\end{minipage}
\caption{Performance comparison across video length categories in Video-MME and LongVideoBench benchmarks.}
\vspace{-0.3cm}
\label{table:longvideo_table}
\end{table}
%%%%%%%%%%%%%%%%%%%%%%%%%%%%%%%%%%%%%%%

\paragraph{Memory-Efficient Setup in MA-GC.} For the MA-GC setup, we aim to minimize memory usage by selecting optimal intervals $l$ and $s$ for layer-wise and sequence-wise checkpoints, respectively, within the bounds $1 \leq l \leq L$ and $1 \leq s \leq S$. These parameters are chosen to reduce the total memory requirement, as expressed in \cref{eq:MemoryEquation}, covering layer and sequence checkpoints and grid checkpoint cells. \cref{alg:ma_gc_forward} manages the forward pass by storing checkpoints $L_{\text{ckpt}}$ and $S_{\text{ckpt}}$ at defined intervals, while \cref{alg:ma_gc_backward} restores these activations during backpropagation, further optimization. For efficiency in SSD \cite{Dao2024TransformersAS}, we restrict $s$ to multiples of 256 when $S \geq 256$, which helps maintain high processing performance in the SSD's state scan logic. This approach, illustrated in \cref{fig:2}, enables memory-efficient handling of long sequence lengths, significantly extending the model's feasible input size during training.

\paragraph{Evaluation Metrics \& Setup.}
We primarily assess our model using two categorized video analysis benchmarks: long video understanding and general video understanding. For long video benchmarks, we employ Video-MME~\cite{fu2024video} and LongVideoBench~\cite{wu2024longvideobench}, both of which include test instances with video durations of up to 2-hours. For shorter, yet more generalized benchmarks, we utilize four different video analysis benchmarks: ActivityNetQA~\cite{yu2019activitynet}, VideoChatGPT~\cite{maaz2023video}, and MVBench~\cite{li2024mvbench}. We used \texttt{gpt-3.5-turbo-0125} to evaluate responses, implementing a termination criterion to handle repetitive sequences, where generation halts if five consecutive tokens previously appeared. Given our primary model, Video-Ma$^2$mba-3.1B, our main comparisons are with baselines that has $\sim$8B parameters.

%%%%%%%%%%%%%%%%%%%%%%%%%%%%%%%%%%%%%%%
\begin{table}[t!]
\centering
\resizebox{1.0\linewidth}{!}{
\begin{tabular}{lccccc}
\Xhline{3\arrayrulewidth}\rule{0pt}{9pt}
&                        & \multicolumn{2}{c}{ActNet-QA} & VCG  & MVBench \\ \cmidrule(lr){3-4} \cmidrule(lr){5-5} \cmidrule(lr){6-6}
\multirow{-2}{*}{Model} & \multirow{-2}{*}{Size} & Acc.                       & Score                     & Acc.  & Acc.     \\ \hline
\rowcolor[HTML]{EFEFEF} 
GPT4V~\cite{gpt4v}                                           & -                                              & 57.0                       & -                         & 4.06 & 43.5    \\
\rowcolor[HTML]{EFEFEF} 
GPT-4o~\cite{gpt4o}                                          & -                                              & 61.9                      & -                         & -    & -       \\
\rowcolor[HTML]{EFEFEF} 
Gemini 1.5 Pro~\cite{reid2024gemini}                                  & -                                              & 57.5                      & -                 & -    & -       \\  \cdashline{1-6}\noalign{\vskip 0.5ex}
VideoLLaMA~\cite{zhang2023video}                                      & 7B                                             & 12.4                      & 1.1                       & 2.16 & 34.1    \\
Video-ChatGPT~\cite{maaz2023video}                                   & 7B                     & 35.2                      & 2.7                       & 2.42 & 32.7    \\
MovieChat~\cite{song2024moviechat}                                       & 7B                     & 45.7                      & -                         & 2.67 & -       \\
Chat-UniVi~\cite{jin2024chat}                                      & 7B                     & 46.1                      & 3.2                       & 2.99 & -       \\
LLaMA-VID~\cite{li2025llama}                                       & 7B                     & 47.4                      & \underline{3.3}                       & 2.89 & 41.3    \\
VideoChat2-Mistral~\cite{li2024mvbench}                              & 7B                     & 49.1                      & \underline{3.3}                       & 2.98 & \textbf{62.3}    \\
ShareGPT4Video~\cite{chen2024sharegpt4video}                                  & 8B                                             & 50.8                      & -                         & -    & 51.2    \\ 
VideoLLaMA2~\cite{cheng2024videollama}                                     & 7B                     & \textbf{53.0}                      & \underline{3.3}                       & \textbf{3.13} & \underline{54.6} \\ \cdashline{1-6}\noalign{\vskip 0.5ex}
\rowcolor[HTML]{ECF4FF} 
Video-Ma$^2$mba-0.7B                   & 0.7B                                           & 43.8                      & 3.2                       & 2.69 & 41.1    \\
\rowcolor[HTML]{ECF4FF}  
Video-Ma$^2$mba-1.8B                   & 1.8B                                           & 50.0                      & 3.1                       & 2.76 & 44.4    \\
\rowcolor[HTML]{ECF4FF} 
Video-Ma$^2$mba-3.1B                   & 3.1B                                           & \underline{51.7}                      & \textbf{3.4}                       & \underline{3.03} & 48.3   \\ \Xhline{3\arrayrulewidth}\rule{0pt}{9pt}
\end{tabular}
}
\vspace{-0.7cm}
\caption{Benchmark results for ActivityNetQA, VideoChatGPT, and MVBench, comparing Video-Ma$^2$mba and baselines.}
\vspace{-0.4cm}
\label{table:general}
\end{table}
%%%%%%%%%%%%%%%%%%%%%%%%%%%%%%%%%%%%%%%

%%%%%%%%%%%%%%%%%%%%%%%%%%%%%%%%%%%%%%%
\begin{table*}[t!]
\centering
\resizebox{0.82\linewidth}{!}{
\begin{tabular}{llcccccccccccc}
\Xhline{3\arrayrulewidth}\rule{0pt}{9pt}
\multirow{2}{*}{Method}                                                & \multirow{2}{*}{Model} & \multicolumn{12}{c}{Sequence Length ($S=2^{n}$)}                                            \\ \cmidrule(lr){3-14} 
                                                                       &                        & 10  & 11  & 12   & 13   & 14   & 15   & 16   & 17   & 18   & 19   & 20   & 21   \\ \hline
\multirow{3}{*}{\begin{tabular}[c]{@{}l@{}}GC off\\ : $O(L \cdot S)$\end{tabular}}  & 350M, $L$=48, $d$=1024     & 0.9 & 1.7 & 3.3  & 6.6  & 13.3 & 26.5 & 52.9 & -    & -    & -    & -    & -    \\
                                                                       & 1.3B, $L$=48, $d$=2048     & 1.7 & 3.3 & 6.5  & 13.1 & 26.0 & 52.1 & -    & -    & -    & -    & -    & -    \\
                                                                       & 2.7B, $L$=64, $d$=2560     & 2.7 & 5.4 & 10.7 & 21.4 & 42.6 & -    & -    & -    & -    & -    & -    & -    \\ \cdashline{1-14}\noalign{\vskip 0.5ex}
\multirow{3}{*}{\begin{tabular}[c]{@{}l@{}}GC on\\ : $O(L \cdot S)$\end{tabular}}   & 350M, $L$=48, $d$=1024     & 0.4 & 0.7 & 1.3  & 2.7  & 5.5  & 10.9 & 21.9 & 43.7 & -    & -    & -    & -    \\
                                                                       & 1.3B, $L$=48, $d$=2048     & 0.7 & 1.4 & 2.7  & 5.5  & 10.9 & 21.8 & 43.5 & -    & -    & -    & -    & -    \\
                                                                       & 2.7B, $L$=64, $d$=2560     & 1.1 & 2.2 & 4.4  & 8.7  & 17.4 & 34.7 & 69.3 & -    & -    & -    & -    & -    \\ \cdashline{1-14}\noalign{\vskip 0.5ex}
\multirow{3}{*}{\begin{tabular}[c]{@{}l@{}}Sqrt GC\\ : $O(\sqrt{L} \cdot S)$\end{tabular}} & 350M, $L$=48, $d$=1024     & 0.2 & 0.4 & 0.8  & 1.6  & 3.1  & 6.2  & 12.3 & 24.6 & 49.3 & -    & -    & -    \\
                                                                       & 1.3B, $L$=48, $d$=2048     & 0.4 & 0.8 & 1.5  & 3.1  & 6.1  & 12.1 & 24.3 & 48.5 & -    & -    & -    & -    \\
                                                                       & 2.7B, $L$=64, $d$=2560     & 0.6 & 1.1  & 2.2  & 4.4  & 8.7  & 17.3 & 34.6 & 69.3 & -    & -    & -    & -    \\ \cdashline{1-14}\noalign{\vskip 0.5ex}
\multirow{3}{*}{\begin{tabular}[c]{@{}l@{}}MA-GC\\ : $O(S)$\end{tabular}}   & 350M, $L$=48, $d$=1024     & 0.3 & 0.5 & 0.6  & 1.1  & 1.6  & 2.4  & 3.8  & 5.5  & 8.8  & 15.4 & 23.1 & 40.2 \\
                                                                       & 1.3B, $L$=48, $d$=2048     & 0.5 & 0.9 & 1.2  & .2.1 & 3.7  & 4.8  & 7.4  & 11.3 & 17.7 & 30.8 & 45.8 & -    \\
                                                                       & 2.7B, $L$=64, $d$=2560     & 0.7 & 1.1 & 1.8  & 2.7  & 4.2  & 6.9  & 10.5 & 17.2 & 25.9 & 42.2 & -    & -    \\ 
                                                                       \Xhline{3\arrayrulewidth}\rule{0pt}{9pt}
\end{tabular}
}
\vspace{-0.7cm}
\caption{Memory overhead (GB) for GC methods in Mamba-2-2.7B across sequence lengths ($S = 2^n$). ``GC off'' indicates no checkpointing; ``GC on'' applies checkpointing per layer; ``Sqrt GC'' groups layers by $\sqrt{L}$; and ``MA-GC'' optimizes based on sequence length. Each cell show peak memory during activation and backpropagation (BF16 precision), excluding model weights and gradients.}
\vspace{-0.4cm}
\label{table:memory}
\end{table*}
%%%%%%%%%%%%%%%%%%%%%%%%%%%%%%%%%%%%%%%

\subsection{Experimental Results}
\paragraph{Results on Long Video Analysis.} 
We report the long video comprehension results using Video-MME~\cite{fu2024video} and LongVideoBench~\cite{wu2024longvideobench} in \cref{table:longvideo_table}. Despite its smaller scale, Video-Ma$^2$mba outperforms most 7B models in both benchmarks. This is due to our method of accessing the entire video at 1-second intervals, contrasting with the sparse frame sampling strategy. Our comprehensive approach facilitates frequent observations, highlighting our framework's effectiveness against larger 7B models. By selectively preserving key information, Video-Ma$^2$mba avoids the typical information loss and computational burdens, delivering precise, context-aware responses for long videos.

\paragraph{Results on General Video Analysis.} In \cref{table:general}, we summarize general video analysis in several benchmarks: ActivityNetQA~\cite{yu2019activitynet}, VideoChatGPT~\cite{maaz2023video}, and MVBench~\cite{li2024mvbench}. These benchmarks are much shorter than the previously used long video benchmarks, but provide a comprehensive foundation for assessing our model's capability to analyze videos and answer related questions. As in the table, we also demonstrate competitive performance of Video-Ma$^2$mba across the benchmarks, holding its own against larger models with parameters over 7B. 

%%%%%%%%%%%%%%%%%%%%%%%%%%%%%%%%%%%%%%%%%%%%%%%%%%%%%%%%%%%%%%
\begin{table}[t!]
\centering
\resizebox{1.0\linewidth}{!}{
\begin{tabular}{ccccccc}
\Xhline{3\arrayrulewidth}\rule{0pt}{9pt}
Tr Stage           & \multicolumn{2}{c}{Frame Limit}  & \multicolumn{4}{c}{Video-MME}                                 \\ \cmidrule(lr){1-1} \cmidrule(lr){2-3} \cmidrule(lr){4-7}      
1/ 1.5 /2          & train                   & infer  & Short: $\leq$2m    & Mid: 4-15m    & Long: 30-60m  & Overall       \\ \hline
\multirow{2}{*}{\Checkmark~~\ding{55}~~\Checkmark} & \multirow{2}{*}{16 frm} & 8 frm  & 49.0          & 38.7          & 33.8          & 40.5          \\
                   &                         & 16 frm & \textbf{50.0} & \textbf{40.7} & \textbf{34.6} & \textbf{41.7} \\ \cdashline{1-7}\noalign{\vskip 0.5ex}
\multirow{4}{*}{\Checkmark~~\ding{55}~~\Checkmark} & \multirow{4}{*}{1 fps}  & 8 frm  & 47.7          & 37.9          & 32.2          & 39.3          \\
                   &                         & 16 frm & 50.6          & 39.4          & 33.2          & 41.1          \\
                   &                         & 32 frm & 52.7          & 40.8          & 33.9          & 42.4    \\
                   &                         & 1 fps  & \textbf{54.4} & \textbf{41.4} & \textbf{34.4} & \textbf{43.4} \\ \cdashline{1-7}\noalign{\vskip 0.5ex}
\multirow{4}{*}{\Checkmark~~\Checkmark~~\Checkmark} & \multirow{4}{*}{1 fps}  & 8 frm  & 53.3          & 39.3          & 32.2          & 41.6          \\
                   &                         & 16 frm & 55.9          & 41.3          & 33.9    & 43.7          \\
                   &                         & 32 frm & \textbf{57.9} & 41.9    & 33.9    & 44.6    \\
                   &                         & 1 fps  & 57.6    & \textbf{42.7} & \textbf{35.4} & \textbf{45.2} \\
\Xhline{3\arrayrulewidth}\rule{0pt}{9pt}
\end{tabular}
}
\vspace{-0.7cm}
\caption{Ablation study on frame size and Stage 1.5 effects in Video-MME using Video-Ma$^2$mba-3.1B.}
\vspace{-0.4cm}
\label{table:ablation}
\end{table}
%%%%%%%%%%%%%%%%%%%%%%%%%%%%%%%%%%%%%%%%%%%%%%%%%%%%%%%%%%%%%%

\vspace{-1mm}
\subsection{Additional Analyses on Video-Ma$^2$mba}
\vspace{-1mm}
\paragraph{Multi-Axis Gradient Checkpointing} For each model configuration, we measured peak memory usage at different sequence lengths to evaluate the memory efficiency of the MA-GC setup. Memory measurements were taken in two steps: (\lowercase\expandafter{\romannumeral1}) capturing baseline memory immediately after loading the model, and (\lowercase\expandafter{\romannumeral2}) recording peak memory during backpropagation. The difference provides the sequence-dependent memory overhead, reflecting the impact of MA-GC on memory savings. Results, shown in \cref{table:memory}, demonstrate a reduction in space complexity from $O(L \cdot S)$ to $O(S)$, highlighting the efficiency gains of MA-GC over standard checkpointing methods.

\paragraph{Effect of Frame Restriction and Stage 1.5.} We conduct ablation studies for the effectiveness of frame threshold and long video knowledge learning (stage 1.5). As summarized in \cref{table:ablation}, when the model was trained with a 16-frame limit, it showed lower performance across all video lengths compared to the our framework, achieving a +1.7 points (4.1\%) improvement. This indicates that limiting frames can cause the model to miss essential visual cues, impacting comprehension. Including stage 1.5 for long video understanding led to consistent gains across all video length, with performance increasing by +1.8 points (4.1\%). Notably, while models trained with 16-frame and 1 FPS limits showed minimal difference without stage 1.5, adding of this stage improved performance with a gain of +1.0 points (2.9\%) due to enhanced context tracking over extended sequences. 
\vspace{-1mm}
\section{Discussion and Conclusion}
\vspace{-1mm}
\paragraph{Discussion.}
Even if we have achieved promising results in our experiments, there are several discussion points and limitations. The recently launched Mamba-2, unlike established Transformer-based LLMs, is still immature with a smaller model size and insufficient QA capabilities from limited language instruction training, potentially capping performance. Future enhancements should expand the architecture and diversify training to fully exploit Mamba-2's potential, possibly outperforming more mature models.

Additionally, we propose a method of feeding entire lengthy frames, up to 0.8M sequence length, all at once to LM backbones, although this is not the only approach to optimizing performance. In particular, for lengthy videos, retrieving targeted salient information through selective frame sampling may prove to be a more effective modeling strategy and remains an intriguing direction for future research.

\paragraph{Conclusion.} In this work, we propose Video-Ma$^2$mba, a novel Large Multi-modal Model designed for efficient long video understanding, which integrates the Mamba structure into the vision modality. In addition, to push the boundary of memory efficiency, we introduce Multi-axis Gradient Checkpointing strategy and achieve significant memory savings, enabling the processing of extended video sequences up to 0.8M context length. Our extensive validation across multiple benchmarks confirms that Video-Ma$^2$mba not only matches but in some cases exceeds the performance of larger models, highlighting the effectiveness and potential of our approach in pushing the boundaries of video sequence modeling.

\newpage
{
    \small
    \bibliographystyle{ieeenat_fullname}
    \bibliography{main}
}

\clearpage
\setcounter{page}{1}
\maketitlesupplementary

\appendix

\section{Interaction Schema}
\label{sec:ModelInteraction}
Video-Ma$^2$mba processes entire video information sequentially at 1 FPS, formulating answers to user queries while retaining relevant details for subsequent questions. Unlike attention-based models, this sequence-modeling approach relies on its ability to remember and retrieve crucial details for future unseen queries. The interaction style is structured in below \cref{table:instruction}.

%%%%%%%%%%%%%%%%%%%%%%%%%%%%%%%%%%%%%%%%%%%%%%%%%
%%%%%%%%%%%%%%%%%%%%%%%%%%%%%%%%%%%%%%%%%%%%%%%%%%%%%%%%%%%%%%%%%
\begin{table}[h!]
\centering
    \begin{minipage}{0.99\columnwidth}\vspace{0mm}    
    \centering
    \begin{tcolorbox}[enhanced, sharp corners, colframe=black, colback=white, boxrule=0.5mm, width=\linewidth, arc=0mm, left=2mm, right=2mm]
        \small
    
        {\color[HTML]{3531FF} \textbf{Video QA Instruction:}}
        
        \raggedright
        \texttt{<|system|>You are a helpful assistant.<|endoftext|>} \\
        \texttt{<|user|>} \\
        \texttt{\quad<frame1\_1><frame1\_2> $\cdots$ <frame1\_144>} \\
        \texttt{\quad<frame2\_1><frame2\_2> $\cdots$ <frame2\_144>} \\
        \texttt{\quad$\cdots$} \\
        \texttt{\quad<frameN\_1><frameN\_2> $\cdots$ <frameN\_144>} \\
        \texttt{({\color[HTML]{3531FF}question\_1})<|endoftext|>} \\
        \texttt{<|assistant|>({\color[HTML]{3531FF}response\_1})<|endoftext|>} \\
        \texttt{<|user|>({\color[HTML]{3531FF}question\_2})<|endoftext|>} \\
        \texttt{<|assistant|>({\color[HTML]{3531FF}response\_2})<|endoftext|>} \\
        \texttt{$\cdots$}
    \end{tcolorbox}
    \vspace{-0.4cm}
    \caption{Illustration of the interaction schema of Video-Ma$^2$mba during Stage 2 of SFT. Frames are processed in sequence to generate responses to user instructions, ensuring continuity across queries.}
    \label{table:instruction}
    \end{minipage}
\end{table}
%%%%%%%%%%%%%%%%%%%%%%%%%%%%%%%%%%%%%%%%%%%%%%%%%%%%%%%%%%%%%%%%%

%%%%%%%%%%%%%%%%%%%%%%%%%%%%%%%%%%%%%%%%%%%%%%%%%

\section{Training Details}
\label{sec:TrainHyperParameters}

\paragraph{Data Implementation for Efficient Training.} To improve the efficiency of training on large-scale video QA datasets due to academic budgets, we deploy a data compression strategy that groups related QA pairs for the same video into a single sample (\textit{i.e.,} single series of QA sets). This drastically reduces the number of training samples from 1.3M to 184K (\textit{e.g.,} LLaVA-Video-178K~\cite{zhang2024video}, NeXT-QA~\cite{xiao2021next}, ActivityNetQA~\cite{yu2019activitynet}, PerceptionTest~\cite{patraucean2024perception}) while preserving the diversity and contextual information of the original dataset. By significantly minimizing decoding overhead, this strategy enables more efficient model training without compromising data diversity.

\paragraph{Training Hyperparameters.}
All Video-Ma$^2$mba variations are trained under consistent configurations, with slight differences in \textit{per-device batch size} due to hardware constraints. To match the \textit{global batch size} across model variations, we utilize \textit{gradient accumulation}, ensuring a similar training schedule for all variations. The hyperparameters for each training stage are detailed in \cref{table:training_hyperparams}.

\paragraph{Batch Size and Gradient Accumulation.}
Each model variation leverages the maximum available GPU memory to determine its per-device batch size. Gradient accumulation is applied to align the global batch size across variations, enabling consistent optimization behavior despite hardware differences.

\paragraph{Training Precision and Gradient Checkpointing.}
We use BFloat16 precision for all stages and adopt \textit{Multi-Axis Gradient Checkpointing}. This approach significantly reduces memory consumption, enabling training with longer sequences and occasionally accommodating larger batch sizes.

\begin{table}[h]
    \centering
    \vspace{-0.1cm}
    \setlength\tabcolsep{10pt}
    \resizebox{1.0\linewidth}{!}{\fontsize{18pt}{23pt}\selectfont
        \begin{tabularx}{\textwidth}{p{4.8cm}|*{3}{>{\centering\arraybackslash}X}}
        config & Stage1 & Stage1.5 & Stage2 \\
        \Xhline{1.0pt}
        input modality           & Vid + Img    & Video              & Video                     \\
        FPS for video            & \multicolumn{3}{c}{1 FPS}                                         \\
        input resolution         & \multicolumn{3}{c}{336x336}                                       \\
        trainable params       & Projector        & Full Model         & Full Model                \\
        LLM lr       & 1e-3             & 4e-5               & 4e-5                      \\
        Vision lr       & -                & 4e-6               & 4e-6                      \\
        lr scheduler   & \multicolumn{3}{c}{Cosine Decay}                                  \\
        optimizer                & \multicolumn{3}{c}{AdamW $(\beta_1=0.9, \beta_2=0.95)$}           \\
        global batch size        & 512              & 32                 & 32                        \\
        train epochs             & 2                & 2                  & 2                         \\
        warmup ratio             & \multicolumn{3}{c}{0.1}                                           \\
        weight decay             & \multicolumn{3}{c}{0.05}                                          \\
        gradient clipping        & \multicolumn{3}{c}{1.0}                                           \\
        training precision       & \multicolumn{3}{c}{BFloat16}                                      \\
        DeepSpeed stage          & \multicolumn{3}{c}{ZeRO-1}                                        \\
        GC                       & \multicolumn{3}{c}{Multi-Axis Gradient Checkpointing}    \\
        \end{tabularx}
    }
    \vspace{-0.3cm}
    \caption{
        Hyperparameters for Training Stages.
    }
    \label{table:training_hyperparams} 
\end{table}
%%%%%%%%%%%%%%%%%%%%%%%%%%%%%%%%%%%%%%%%%%%%%%%%%%%%%%%%%%%%%%

\section{Memory Estimation Logic}
\label{sec:appendix_memory_estimation}

The memory estimation logic in Video-Ma$^2$mba optimizes memory requirements by determining the ideal checkpointing intervals $l$ and $s$ used during forward and backward passes. The constants $C_{S\text{-ckpt}}$, $C_{L\text{-ckpt}}$, $C_\text{grid}$, and $C_\text{state}$ depend on the backbone model's configuration, including SSM implementation, block design, and precision type.

Accordingly, the total expected memory $M$ can be computed as follows:
%%%%%%%%%%%%%%%%%%%%%%%%%%%%%%%%%%%%%%%%%%%%%%%%%%%%%%%%%%%%%%
\begin{equation}
    M = M_{L\text{-ckpt}} + M_{S\text{-ckpt}} + M_{\text{grid}} + M_{\text{state}},
\end{equation}
%%%%%%%%%%%%%%%%%%%%%%%%%%%%%%%%%%%%%%%%%%%%%%%%%%%%%%%%%%%%%%
where the each memory component in the above equation are defined as:
%%%%%%%%%%%%%%%%%%%%%%%%%%%%%%%%%%%%%%%%%%%%%%%%%%%%%%%%%%%%%%
\begin{align}
    M_{L\text{-ckpt}} &= \frac{L S}{l} \cdot C_{L\text{-ckpt}}, \\
    M_{S\text{-ckpt}} &= \frac{L S}{s} \cdot C_{S\text{-ckpt}}, \\
    M_{\text{grid}} &= l s \cdot C_{\text{grid}}, \\
    M_{\text{state}} &= s \cdot C_{\text{state}}.
\end{align}
%%%%%%%%%%%%%%%%%%%%%%%%%%%%%%%%%%%%%%%%%%%%%%%%%%%%%%%%%%%%%%

\subsection{Model-Specific Constants}
\cref{table:model_constants} outlines the constants for three backbone configurations of Mamba-2 in the BFloat16 precision setting. Note that SSM states use Float32 precision, affecting $C_\text{state}$ and $C_{S\text{-ckpt}}$ depending on whether half- or single-precision is used.

%%%%%%%%%%%%%%%%%%%%%%%%%%%%%%%%%%%%%%%%%%%%%%%%%%%%%%%%%%%%%%
\begin{table}[h!]
    \centering
    \resizebox{\linewidth}{!}{
        \begin{tabular}{lcccc}
            \Xhline{3\arrayrulewidth} \rule{0pt}{9pt}
            Model       & $C_{L\text{-ckpt}}$ & $C_{S\text{-ckpt}}$ & $C_{\text{grid}}$ & $C_{\text{state}}$ \\ \hline
            Mamba-2-370m & 1,024      & 269,056    & 6,432     & 264,448    \\
            Mamba-2-1.3b & 2,048      & 537,344    & 12,608    & 528,640    \\
            Mamba-2-2.7b & 2,560      & 671,488    & 15,696    & 660,736    \\ 
            \Xhline{3\arrayrulewidth} \rule{0pt}{9pt}
        \end{tabular}
    }
    \vspace{-7.5mm}
    \caption{Model-specific constants for memory estimation under BFloat16 precision. Constants reflect relative element counts, where SSM states in Float32 are equivalent to two BFloat16 elements.}
    \label{table:model_constants}        
\end{table}
%%%%%%%%%%%%%%%%%%%%%%%%%%%%%%%%%%%%%%%%%%%%%%%%%%%%%%%%%%%%%%

\paragraph{Recomputation for Memory Optimization.}
The term $M_\text{state}$, omitted in \cref{eq:MemoryEquation}, arises in Mamba-2~\cite{Dao2024TransformersAS} due to recomputation of SSM states during the backward pass. This recomputation reduces memory usage by avoiding the storage of intermediate states during forward computation:

%%%%%%%%%%%%%%%%%%%%%%%%%%%%%%%%%%%%%%%%%%%%%%%%%%%%%%%%%%%%%%
\begin{quote}
    \textit{``The intermediate states are not stored but recomputed in the backward pass when the inputs are loaded from HBM to SRAM. As a result, the fused selective scan layer has the same memory requirements as an optimized transformer implementation with FlashAttention.''}~\cite{Gu2023MambaLS}
\end{quote}
%%%%%%%%%%%%%%%%%%%%%%%%%%%%%%%%%%%%%%%%%%%%%%%%%%%%%%%%%%%%%%

Although $M_{\text{state}}$ grows linearly with $s$ (\textit{i.e.,} $M_{\text{state}}{=}O(s)$), the memory term $M_{\text{grid}}$ grows as the product of $l \cdot s$ (\textit{i.e.,} $M_{\text{grid}}{=}O(l s)$), making it asymptotically dominant. Consequently, $M_{\text{state}}$ becomes a negligible term in the overall memory complexity, and the analysis in \cref{sec:analysis} remains valid.

\paragraph{Checkpointing Trade-Offs.} Selecting the optimal values for $l$ and $s$ is critical to minimizing $M$. For example, larger values of $s$ reduce $M_{S\text{-ckpt}}$ but increase restoration overhead during the backward pass ($M_\text{grid}$ and $M_\text{state}$). As shown in \cref{table:model_constants}, a careful balance for memory savings is required to process long video sequences efficiently.

\section{Gradient Checkpointing Time Analysis}
\label{sec:GCTimes}
We analyze the computational efficiency of various gradient checkpointing methods, with results summarized in \cref{table:TimeSummary}. Throughput (tokens per second) indicates the processing speed, while per-token processing time (milliseconds per token) provides a more detailed perspective on computational overhead. The reported measurements represent the median of six runs conducted on an A100 80GB GPU using the Mamba-2-2.7b, which is the backbone of Video-Ma$^2$mba-3.1B. A warm-start configuration was used to minimize initialization overhead, and the times include both the forward and backward computation steps.

MA-GC demonstrates the ability to train on sequence lengths up to \textbf{32$\times$ longer} than the GC-off baseline within the same memory constraints. Although this extended capability comes with a \textbf{35\% reduction} in throughput, this is a trade-off that enables scalable and efficient training for tasks requiring extremely long sequences.

The \textit{gradient checkpointing} methods can save memory and handle longer sequences but are fundamentally constrained when processing extremely long sequences due to memory limitations. MA-GC overcomes such issues by introducing a multi-axis gradient checkpointing mechanism, enabling training on sequence lengths up to 32$\times$ longer while retaining computational feasibility. By efficiently managing memory without incurring prohibitive overhead, MA-GC balances performance and resource efficiency, significantly extending the scalability of large-scale models.

%%%%%%%%%%%%%%%%%%%%%%%%%%%%%%%%%%%%%%%%%%%%%%%%%%%%%%%%%%%%%%
\begin{table}[t]
    \centering
    \resizebox{\linewidth}{!}{
        \begin{tabular}{lcc}
            \Xhline{3\arrayrulewidth} \rule{0pt}{9pt}
            & \textbf{Throughput} $\uparrow$ & \textbf{Processing Time} $\downarrow$ \\
            \textbf{Method} & \textbf{(tokens/s)} & \textbf{(ms/token)} \\ \hline
            GC off @ $2^{14}$ & 12,167.86 & 0.082 \\
            GC on @ $2^{16}$ & 8,449.93 & 0.118 \\
            Sqrt GC @ $2^{17}$ & 8,617.66 & 0.116 \\
            MA-GC @ $2^{19}$ & 7,913.58 & 0.126 \\
            \Xhline{3\arrayrulewidth} \rule{0pt}{9pt}
        \end{tabular}
    }
    \vspace{-6.5mm}
    \caption{Computational analysis of throughput and per-token processing time among gradient checkpointing methods. Results are measured using the Mamba-2-2.7b model on an A100 80GB GPU. The notation \textbf{{\boldmath$@$} $\mathbf{2^n}$} specifies the sequence length (in tokens) used for measurement.}
    \label{table:TimeSummary}        
\end{table}

%%%%%%%%%%%%%%%%%%%%%%%%%%%%%%%%%%%%%%%%%%%%%%%%%%%%%%%%%%%%%%

\section{Qualitative Evaluation}
\label{sec:qeval} 

\begin{figure*}[t!]
\centering
\includegraphics[width=1.0\linewidth]{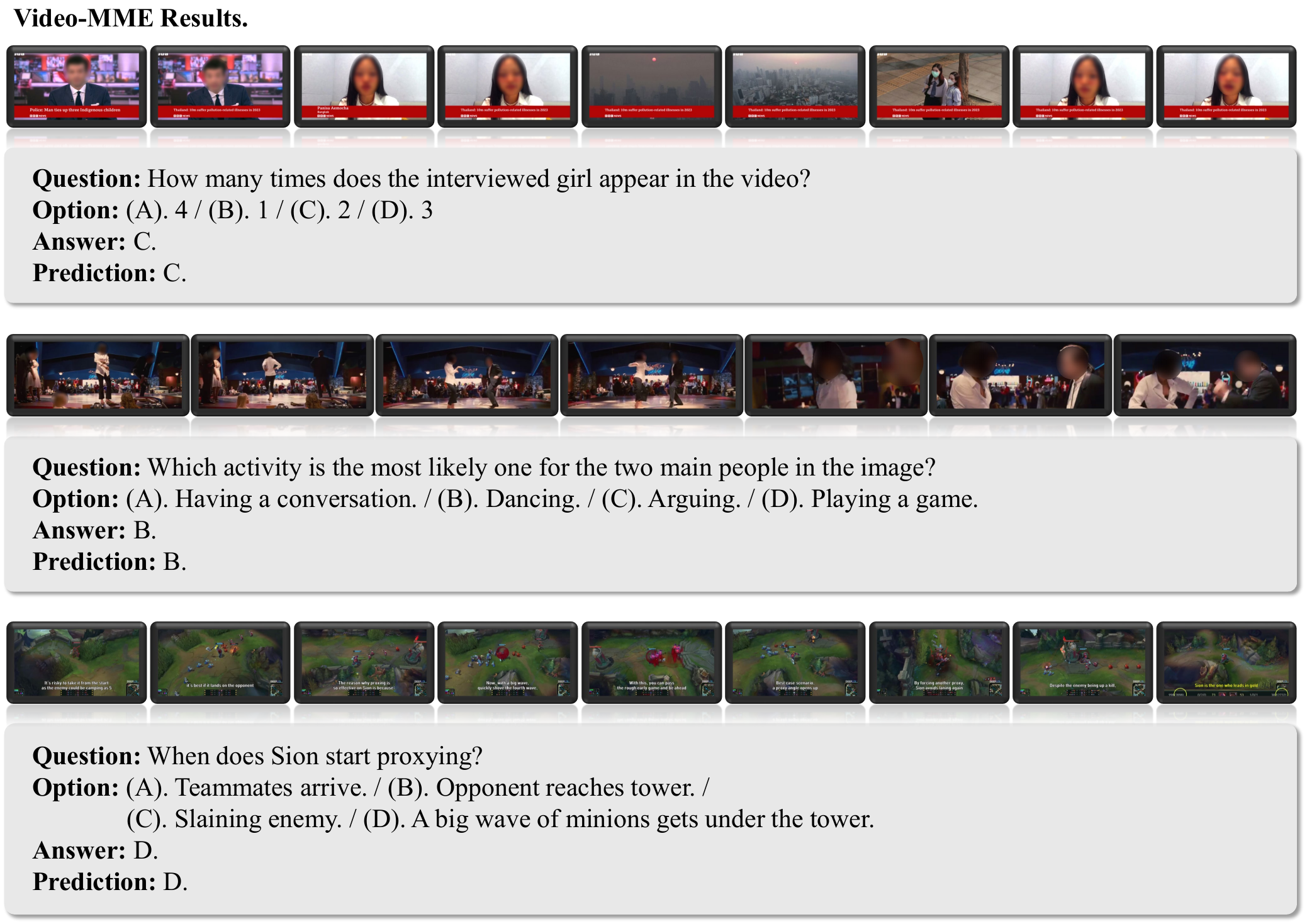}
\vspace*{-0.8cm}
\caption{Qualitative examples on Video-MME~\cite{fu2024video} with Video-Ma$^2$mba-3.1B.}
\vspace*{-0.5cm}
\label{fig:4}
\end{figure*}

\begin{figure*}[t!]
\centering
\includegraphics[width=1.0\linewidth]{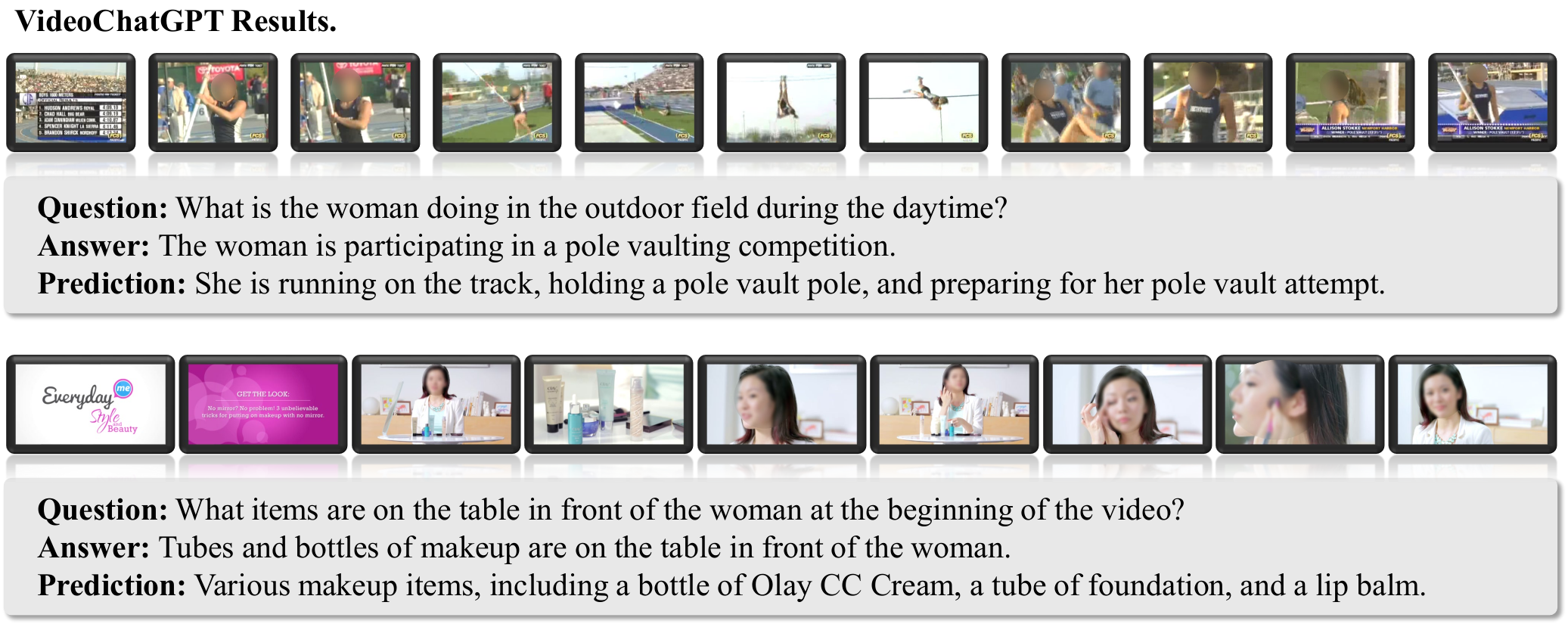}
\vspace*{-0.8cm}
\caption{Qualitative examples from the Generative Subset of VideoChatGPT~\cite{maaz2023video} with Video-Ma$^2$mba-3.1B.}
\vspace*{-0.5cm}
\label{fig:5}
\end{figure*}

\begin{figure*}[t!]
\centering
\includegraphics[width=1.0\linewidth]{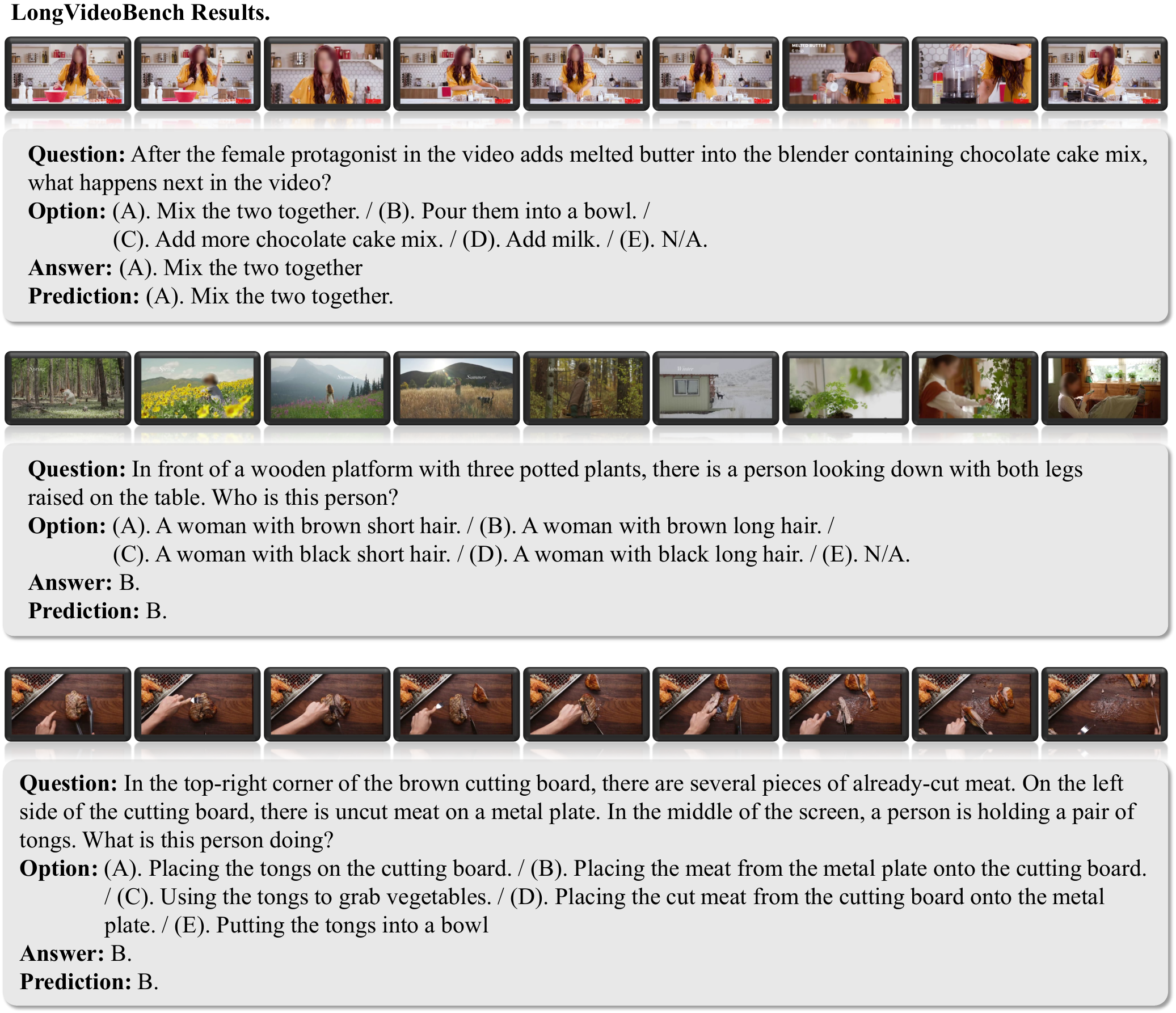}
\vspace*{-0.8cm}
\caption{Qualitative examples on LongVideoBench~\cite{wu2024longvideobench} with Video-Ma$^2$mba-3.1B.}
\vspace*{-0.5cm}
\label{fig:6}
\end{figure*}

As in \cref{fig:4}, \cref{fig:5}, and \cref{fig:6}, our analysis provides qualitative outcomes across various benchmarks, clearly demonstrating the adaptability and effectiveness of Video-Ma$^2$mba. Consistent performance across different benchmarks highlights the model's ability to efficiently handle and analyze disparate datasets. This effectiveness stems from Video-Ma$^2$mba's ability to process extensive context using the MA-GC mechanism, which allows it to handle all incoming inputs comprehensively, facilitating robust data processing and insightful reasoning.

%################################################################################
% Figure
% %################################################################################

% %################################################################################
% Figure
% %################################################################################

% %################################################################################
% Figure
% %################################################################################

\end{document}